\documentclass[sn-mathphys-num]{sn-jnl}

\usepackage{graphicx}%
\usepackage{multirow}%
\usepackage{amsmath,amssymb,amsfonts}%
\usepackage{amsthm}%
\usepackage[section]{placeins}%
\usepackage{mathrsfs}%
\usepackage[title]{appendix}%
\usepackage{xcolor}%
\usepackage{textcomp}%
\usepackage{manyfoot}%
\usepackage{booktabs}%
\usepackage{algorithm}%
\usepackage{algorithmicx}%
\usepackage{algpseudocode}%
\usepackage{listings}%
\usepackage{float}%
\restylefloat{table}%
\usepackage{tabularray}%

\theoremstyle{thmstyleone}%
\theoremstyle{thmstyletwo}%

\theoremstyle{thmstylethree}%

\begin{document}

\title[Article Title]{Can artificial intelligence predict clinical trial outcomes?}


\author[1]{\fnm{Shuyi} \sur{Jin}}

\author[3]{\fnm{Lu} \sur{Chen}}

\author[2]{\fnm{Rongru} \sur{Ding}}

\author[2]{\fnm{Meijie} \sur{Wang}}

\author*[2]{\fnm{Lun} \sur{Yu}}\email{lunyu@metanovas.com}

\affil[1]{\orgdiv{Department of Biomedical Informatics}, \orgname{National University of Singapore}, \orgaddress{\postcode{119077}, \country{Singpaore}}}

\affil*[2]{\orgname{Metanovas Biotech}, \orgaddress{\city{San Francisco}, \postcode{94108}, \state{CA}, \country{USA}}}

\affil[3]{\orgdiv{Department of Material Science}, \orgname{Southeast University}, \orgaddress{\city{Nanjing}, \postcode{210018}, \state{Jiangsu}, \country{China}}}


\abstract{This study evaluates the performance of large language models (LLMs) and the HINT model in predicting clinical trial outcomes, focusing on metrics including Balanced Accuracy, Matthews Correlation Coefficient (MCC), Recall, and Specificity. Results show that GPT-4o achieves superior overall performance among LLMs but, like its counterparts (GPT-3.5, GPT-4mini, Llama3), struggles with identifying negative outcomes. In contrast, HINT excels in negative sample recognition and demonstrates resilience to external factors (e.g., recruitment challenges) but underperforms in oncology trials, a major dataset component. LLMs exhibit strengths in early-phase trials and simpler endpoints like Overall Survival (OS), while HINT shows consistency across trial phases and excels in complex endpoints (e.g., Objective Response Rate). Trial duration analysis reveals improved model performance for medium-to-long-term trials, with GPT-4o and HINT displaying stability and enhanced specificity, respectively. We underscores the complementary potential of LLMs (e.g., GPT-4o, Llama3) and HINT, advocating for hybrid approaches to leverage GPT-4o’s predictive power and HINT’s specificity in clinical trial outcome forecasting.}




\maketitle

\section{Introduction}\label{sec1}
The global clinical trial market has shown continuous growth in recent years. By 2030, the global clinical trial market size is expected to reach 78.3 billion USD\cite{bib1}, with a compound annual growth rate (CAGR) of 5.8\% during the forecast period\cite{bib2}. However, the complexity and cost of clinical trials present significant challenges for companies and research institutions. Advances in drug development, particularly the rise of gene therapies and biologics, have heightened the complexity of clinical trials, leading to greater difficulties in trial design, patient recruitment, and data management\cite{bib3}. In addition, the requirement for large patient populations, especially in Phase III clinical trials, coupled with extended trial durations and escalating costs, exacerbates these challenges\cite{bib4}. Studies have shown that the median cost of Phase I clinical trials is \$3.4 million, Phase II trials cost \$8.6 million, and Phase III trials escalate to \$21.4 million\cite{bib5}. Furthermore, the median duration of non-oncology clinical trials ranges from 5.9 to 7.2 years, while oncology trials have a significantly longer median duration of 13.1 years\cite{bib6}. To address these issues, an increasing number of research institutions and companies have begun exploring the potential of artificial intelligence in accelerating drug development, optimizing trial design, and reducing costs. 
\newline
Artificial Intelligence in healthcare is rapidly expanding, and one emerging area where AI is making an impact is in predicting clinical trial outcome prediction tasks. In recent years, many researchers have employed deep learning methods to predict clinical trial outcomes. For example, Schperberg et al. used a random forest model to predict oncology outcomes in randomized clinical trials\cite{bib7}. Similarly, Qaiser et al. employed a weakly supervised survival convolutional neural network (WSS-CNN) with a visual attention mechanism to predict overall survival rates\cite{bib8}. These studies highlight the potential of artificial intelligence in forecasting clinical trial results. Notably, Fu et al. developed the HINT model, a hierarchical interaction network model that leverages multimodal data such as drug properties, disease information, and trial eligibility criteria to generate embedding vectors\cite{bib9}. This model employs a dynamic attention-based graph neural network to capture the interactive effects among various trial elements, aiming to predict trial success. 
\newline
However, these models still have certain limitations. Most are tailored to specific disease categories or clinical trial endpoints, which restricts their applicability and limits their ability to handle diverse and complex data types. Additionally, previous studies often require standardized data inputs and cannot automatically predict clinical trial outcomes directly from protocol documents. Large language models, with their advanced natural language processing capabilities, may address these challenges by processing unstructured text and providing more generalized predictive capabilities.
\newline
Large language models (LLMs), with their advanced natural language understanding capabilities, offer flexibility and generalizability in various applications\cite{bib10}. By training on extensive clinical data and literature, LLMs can uncover latent patterns and complex relationships, demonstrating unique advantages in the decision-making process of drug development. Previous studies have applied large language models to clinical trial prediction tasks with promising results. For instance, Reinisch et al. used LLMs to predict clinical trial phase transitions\cite{bib11}. Lai et al. evaluated the risk of bias in randomized clinical trials using LLMs\cite{bib12}. Jin et al. employed LLMs to match patients to appropriate clinical trials\cite{bib13}, and Markey et al. utilized LLMs to assist in writing clinical trial documents\cite{bib14}. These examples illustrate that LLMs exhibit a high degree of adaptability in handling diverse trial data and flexible data formats, showing promising potential, particularly in the complex field of trial outcome prediction. Therefore, we considered whether LLMs could be used to predict the likelihood of clinical trial success, enabling us to anticipate trial outcomes and address potential issues that may lead to trial failure\cite{bib15}. 
\newline
In this work, we evaluated the capabilities of mainstream LLMs and the HINT model in predicting clinical trial outcomes. We curated and annotated a dataset of clinical trial information based on the Clinical Trials Database\cite{bib16} and employed these models to perform the predictive task. Subsequently, we assessed the model performance using various metrics to demonstrate the application potential of LLMs in this context. Our analysis aims to offer new insights for risk prediction and trial design in future drug development.
\FloatBarrier
\section{Results}\label{sec2}
\subsection{Prediction on the Entire Clinical Trial Dataset}\label{subsec3}
The test results across all models on the complete dataset are shown in the Table \ref{tab1}. We observed that the GPT-4o model performed the best in terms of balanced accuracy and MCC, reaching a balanced accuracy of 0.573 and an MCC of 0.212(Table \ref{tab1}). Although its recall was high at 0.931, its specificity remained relatively low at 0.214, suggesting a tendency to over-classify trials as successful while overlooking some potential failed trials.
\newline
GPT-4mini and GPT-4, while achieving perfect recall (1), exhibited extremely low specificity (0 and 0.059, respectively), indicating a strong bias toward predicting all cases as positive, resulting in balanced accuracies of 0.500 and 0.529. GPT-3.5 also demonstrated a similar trend with a recall of 0.998 and a specificity of 0.008, leading to a balanced accuracy of 0.503. This likely suggests that these three models have weak predictive power for the trial results across the overall dataset
\newline
The Llama3 model achieved moderate recall (0.964) but low specificity (0.092), resulting in a balanced accuracy of 0.528 and a low MCC of 0.116, indicating limited effectiveness in correctly identifying negative samples. However, it outperformed GPT-4, GPT-4mini, and GPT-3.5
\newline
Among all models, the HINT model achieved the highest specificity at 0.473, which contributed to its relatively strong balanced accuracy of 0.525 and an MCC of 0.050. Despite having lower recall and balanced accuracy compared to other models, HINT's ability to identify negative cases suggests a more balanced performance across positive and negative samples.
\newline
\begin{table}[h]
\centering
\caption{Model Performance Metrics}\label{tab1}
\begin{tabular}{@{}llllll@{}}
\toprule
\textbf{Model} & \textbf{Balanced\_acc} & \textbf{MCC} & \textbf{Recall} & \textbf{Specificity} \\ 
\midrule
GPT-4mini & 0.500 & 0.000 & 1.000 & 0.000 \\ 
GPT-4o    & 0.532 & 0.098 & 0.913 & 0.150 \\ 
GPT-4     & 0.529 & 0.171 & 1.000 & 0.059 \\ 
GPT-3.5   & 0.503 & 0.046 & 0.998 & 0.008 \\ 
HINT     & 0.525 & 0.050 & 0.577 & 0.473 \\ 
Llama3   & 0.528 & 0.116 & 0.964 & 0.092 \\ 
\bottomrule
\end{tabular}
\end{table}
\FloatBarrier
\subsection{Prediction of Different Clinical Trial Phases}\label{subsec3}
As clinical trials progress from Phase I to Phase III, different models exhibit varying trends in metrics such as balanced accuracy, Matthews correlation coefficient (MCC), and specificity. For example, HINT’s metrics improve as the trial phase progresses, while Llama3 demonstrates stronger performance in predicting earlier trial phases.
\newline
In Phase I (Table \ref{tab2}), GPT-4o and GPT-4 achieve relatively high balanced accuracies of 0.557 and 0.600, respectively, but their specificities are low, at 0.40 and 0.20. This indicates certain limitations in identifying negative samples. Llama3 also demonstrates a relatively strong balanced accuracy of 0.556, though accompanied by a lower specificity. GPT-4mini shows poor performance across various metrics, indicating a lack of predictive power for clinical trials. HINT achieves a balanced accuracy of 0.470 and an MCC of -0.058 in Phase I, while GPT-3.5 scores 0.497 and -0.053, suggesting that both models have predictive capabilities weaker than random guessing at this stage. However, it is worth noting that HINT maintains the highest specificity of 0.412, indicating its continued strong ability to identify positive samples at this point.
\newline
In Phase II, the performance of GPT-4o declines, with its balanced accuracy dropping to 0.533, specificity decreasing to 0.067, although MCC shows a slight increase. The balanced accuracy of GPT-4 drops to 0.54, while its MCC decreases to 0.199 and specificity further decreases to 0.008. This downward trend is also observed in the Llama3 model. On the other hand, the HINT model's balanced accuracy increases to 0.517, with improvements in all other metrics. Compared to Phase I, HINT shows improved predictive ability in Phase II clinical trials. GPT-3.5 also shows a slight improvement but does not demonstrate strong predictive power for clinical trials.
\newline
By Phase III, GPT-4o achieves its best performance, with balanced accuracy increasing to 0.625 and MCC reaching 0.408. Its specificity also improves to 0.25, continuing to display relatively stable performance. The HINT model further enhances its performance in this phase, achieving a balanced accuracy of 0.593 and an increase in specificity to 0.473. This demonstrates a balanced capability in recognizing both positive and negative samples. GPT-3.5 and Llama3 exhibit low specificities and weak abilities in identifying negative samples, while GPT-4mini's specificity remains at 0 across all phases.
\newline
\FloatBarrier
\begin{table}[h]
\centering
\caption{Model Performance Metrics}\label{tab2}
\begin{tabular}{@{}llllllll@{}}
\toprule
\textbf{Phase} & \textbf{Model} & \textbf{Balanced\_acc} & \textbf{MCC} & \textbf{Recall} & \textbf{Specificity} \\ 
\midrule
\multicolumn{6}{c}{\textbf{Phase 1 Results}} \\ 
1 & GPT-4mini & 0.500 & 0.000 & 1.000 & 0.000 \\ 
2 & GPT-4o    & 0.557 & 0.120 & 0.714 & 0.400 \\ 
3 & GPT-4     & 0.600 & 0.333 & 1.000 & 0.200 \\ 
4 & GPT-3.5   & 0.497 & -0.053 & 0.995 & 0.000 \\ 
5 & HINT     & 0.470 & -0.058 & 0.528 & 0.412 \\ 
6 & Llama3   & 0.556 & 0.185 & 0.946 & 0.167 \\ 
\midrule
\multicolumn{6}{c}{\textbf{Phase 2 Results}} \\ 
1 & GPT-4mini & 0.500 & 0.000 & 1.000 & 0.000 \\ 
2 & GPT-4o    & 0.533 & 0.175 & 1.000 & 0.067 \\ 
3 & GPT-4     & 0.542 & 0.199 & 1.000 & 0.083 \\ 
4 & GPT-3.5   & 0.504 & 0.051 & 0.998 & 0.009 \\ 
5 & HINT     & 0.518 & 0.036 & 0.508 & 0.528 \\ 
6 & Llama3   & 0.537 & 0.146 & 0.965 & 0.110 \\ 
\midrule
\multicolumn{6}{c}{\textbf{Phase 3 Results}} \\ 
1 & GPT-4mini & 0.500 & 0.000 & 1.000 & 0.000 \\ 
2 & GPT-4o    & 0.625 & 0.408 & 1.000 & 0.250 \\ 
3 & GPT-4     & 0.500 & 0.000 & 1.000 & 0.000 \\ 
4 & GPT-3.5   & 0.505 & 0.060 & 0.998 & 0.012 \\ 
5 & HINT     & 0.594 & 0.193 & 0.714 & 0.474 \\ 
6 & Llama3   & 0.518 & 0.084 & 0.973 & 0.062 \\ 
\bottomrule
\end{tabular}
\end{table}
\FloatBarrier
\subsection{Prediction on Clinical Trials Across Different Disease Categories}\label{subsec3}
\begin{figure}[h]
\centering
\includegraphics[width=0.9\textwidth]{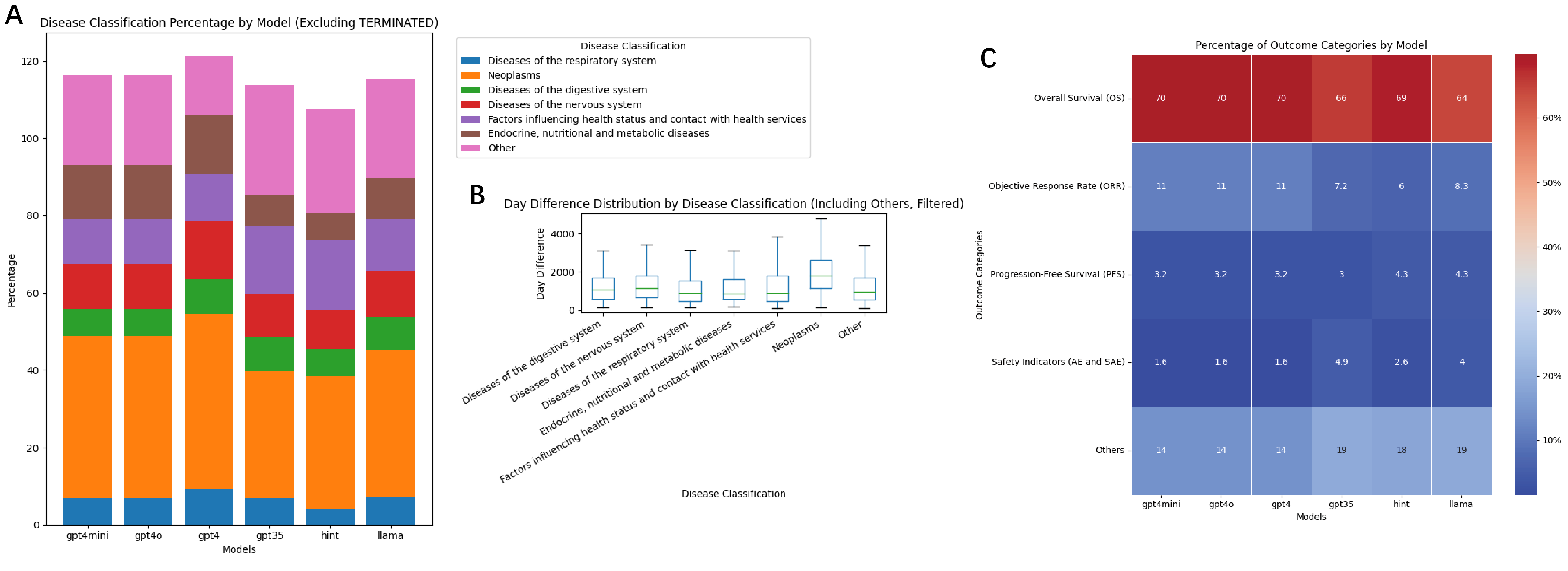}
\caption{A: The distribution of diseases across different model datasets. B:Day difference distribution by difference classification C: Distribution of clinical trial outcomes across different model datasets.}\label{fig1}
\end{figure}

The distribution of diseases across different model datasets is shown in Fig.\ref{fig1} A. Neoplasms consistently comprise the largest portion of the dataset(30\%-37\%), while respiratory diseases have the smallest representation(5\%-9\%). Other categories, such as digestive system diseases, nervous system diseases, and endocrine/metabolic diseases, are also present in notable proportions across all models. Additionally, the "Other" category (for example: Diseases of the eye and adnexa, Diseases of the ear and mastoid process, Diseases of the skin and subcutaneous tissue) appears prominently in the overall distribution. The models exhibit distinct strengths and limitations when predicting clinical trials across different disease categories.For example, the HINT model achieved high average accuracy in "Diseases of the respiratory system" and "Diseases of the digestive system," reaching about 0.75, while the large language models showed lower accuracy for these two categories, around 0.5.
\newline
The HINT model shows significantly lower performance in oncology trials, with a low balanced accuracy of 0.437 and an MCC of -0.137, suggesting that the high complexity and challenging nature of these trials negatively impacted the model's performance. In contrast to previous observations, large language models did not exhibit such a disparity. Most large language models showed predictive abilities in oncology trials that were similar to their overall average performance.
\newline
\newline
Regarding trial complexity, the clinical trial complexity score developed by Markey et al. ranks oncology trials as the most complex across all categories\cite{bib14}. Additionally, the Tufts Center for the Study of Drug Development (CSDD) reports that oncology trials surpass other fields in both complexity and duration\cite{bib17}. In our study, statistical analysis revealed that oncology trials have an average duration approximately 1,000 days longer than other trials. Even trials classified as short-term within oncology would still be considered relatively long compared to trials in other categories (Fig.\ref{fig1} B).
\newline
In terms of recall, large language models consistently achieved high rates, indicating a strong capability in identifying positive cases across various disease categories. However, the HINT model showed a low recall rate of 0.246 in oncology-related diseases, reflecting its limited ability to recognize positive cases in this challenging category. Conversely, the situation was the opposite in other disease categories; for diseases such as endocrine, nutritional, and metabolic disorders, the HINT model achieved a perfect recall rate of 1, indicating a stronger capacity for identifying positive cases within these categories.
\newline
Despite the limited number of experiments, GPT-4o displayed strong recognition of negative samples, especially in respiratory and digestive diseases, where it identified many negative cases, achieving specific rates of 1 and 0.5, far higher than other categories. The Llama3 model, on the other hand, showed poor predictive performance in clinical trials for endocrine, nutritional, and metabolic diseases, as well as diseases of the digestive system, with balanced accuracy lower than random chance and almost no ability to identify negative samples in these categories.However, in other disease classifications, Llama3 has shown relatively excellent predictive ability, especially noteworthy is that: Llama3 demonstrated the strongest predictive ability for Neoplasms among all models, with an average accuracy of 0.555 and an MCC of 0.233. Other models, like GPT-3.5 and GPT-4mini, performed considerably worse, showing almost no detection of negative samples. Although the HINT model’s specificity was not as high as GPT-4o’s, it remained stable in respiratory and non-pathological diseases
\newline
\FloatBarrier
\begin{table}[h]
\centering
\caption{Disease Category Performance Metrics by Model}\label{tab3}
\resizebox{\textwidth}{!}{
\begin{tabular}{@{}llllllll@{}}
\toprule
\textbf{Disease Category} & \textbf{Trails Count} & \textbf{Balanced\_acc} & \textbf{MCC} & \textbf{Recall} & \textbf{Specificity} \\ 
\midrule
\multicolumn{6}{c}{\textbf{Model: GPT-4mini}} \\ 
Diseases of the respiratory system & 3 & 0.500 & 0.0 & 1.000 & 0.0 \\ 
Endocrine, nutritional and metabolic diseases & 6 & 0.500 & 0.0 & 1.000 & 0.0 \\ 
Neoplasms & 18 & 0.500 & 0.0 & 1.000 & 0.0 \\ 
Diseases of the digestive system & 3 & 0.500 & 0.0 & 0.000 & 0.0 \\ 
Diseases of the nervous system & 5 & 0.500 & 0.0 & 1.000 & 0.0 \\ 
Non-pathological diseases & 6 & 0.500 & 0.0 & 1.000 & 0.0 \\ 
Other & 2 & 0.500 & 0.0 & 1.000 & 0.0 \\ 
\midrule
\multicolumn{6}{c}{\textbf{Model: GPT-4o}} \\ 
Diseases of the respiratory system & 3 & 0.750 & 0.500 & 0.500 & 1.000 \\ 
Endocrine, nutritional and metabolic diseases & 6 & 0.250 & -0.632 & 0.500 & 0.000 \\ 
Neoplasms & 18 & 0.512 & 0.040 & 0.900 & 0.125 \\ 
Diseases of the digestive system & 3 & 0.750 & 0.500 & 1.000 & 0.500 \\ 
Diseases of the nervous system & 5 & 0.200 & 0.000 & 0.000 & 0.200 \\ 
Non-pathological diseases & 6 & 1.000 & 1.000 & 1.000 & 1.000 \\ 
Other & 2 & 0.750 & 0.638 & 1.000 & 0.500 \\ 
\midrule
\multicolumn{6}{c}{\textbf{Model: GPT-4}} \\ 
Diseases of the respiratory system & 3 & 0.750 & 0.500 & 0.500 & 1.000 \\ 
Endocrine, nutritional and metabolic diseases & 5 & 0.500 & 0.00 & 1.000 & 0.000 \\ 
Neoplasms & 15 & 0.500 & 0.00 & 1.000 & 0.000 \\ 
Diseases of the digestive system & 3 & 0.750 & 0.500 & 0.100 & 0.500 \\ 
Diseases of the nervous system & 5 & 0.200 & 0.000 & 0.000 & 0.200 \\ 
Non-pathological diseases & 5 & 0.500 & 0.000 & 1.000 & 0.000 \\ 
Other & 2 & 0.786 & 0.670 & 1.000 & 0.571 \\ 
\midrule
\multicolumn{6}{c}{\textbf{Model: GPT-3.5}} \\ 
Diseases of the respiratory system & 148 & 0.501 & 0.011 & 0.988 & 0.015 \\ 
Endocrine, nutritional and metabolic diseases & 173 & 0.500 & 0.000 & 1.000 & 0.000 \\ 
Neoplasms & 712 & 0.504 & 0.064 & 1.000 & 0.008 \\ 
Diseases of the digestive system & 192 & 0.525 & 0.155 & 0.994 & 0.056 \\ 
Diseases of the nervous system & 241 & 0.504 & 0.061 & 1.000 & 0.008 \\ 
Non-pathological diseases & 410 & 0.497 & -0.038 & 0.992 & 0.001 \\ 
Other & 287 & 0.506 & 0.057 & 0.996 & 0.015 \\ 
\midrule
\multicolumn{6}{c}{\textbf{Model: HINT}} \\ 
Diseases of the respiratory system & 15 & 0.732 & 0.464 & 0.750 & 0.714 \\ 
Endocrine, nutritional and metabolic diseases & 27 & 0.636 & 0.426 & 1.000 & 0.273 \\ 
Neoplasms & 132 & 0.437 & -0.137 & 0.246 & 0.627 \\ 
Diseases of the digestive system & 27 & 0.670 & 0.346 & 0.769 & 0.571 \\ 
Diseases of the nervous system & 38 & 0.550 & 0.100 & 0.600 & 0.500 \\ 
Non-pathological diseases & 74 & 0.576 & 0.130 & 0.595 & 0.556 \\ 
Other & 70 & 0.575 & 0.117 & 0.718 & 0.431 \\ 
\midrule
\multicolumn{6}{c}{\textbf{Model: Llama3}} \\ 
Diseases of the respiratory system & 23 & 0.575 & 0.219 & 0.929 & 0.222 \\ 
Endocrine, nutritional and metabolic diseases & 35 & 0.479 & -0.116 & 0.958 & 0.000 \\ 
Neoplasms & 124 & 0.555 & 0.223 & 0.985 & 0.125 \\ 
Diseases of the digestive system & 27 & 0.495 & -0.015 & 0.846 & 0.143 \\ 
Diseases of the nervous system & 39 & 0.554 & 0.178 & 0.950 & 0.158 \\ 
Non-pathological diseases & 46 & 0.562 & 0.302 & 1.000 & 0.125 \\ 
Other & 324 & 0.533 & 0.108 & 0.947 & 0.119 \\ 
\bottomrule
\end{tabular}}
\end{table}
\FloatBarrier
\subsection{Prediction for Different Clinical Endpoints Trials}\label{subsec3}
The Fig.\ref{fig1} C illustrates the distribution of clinical trial outcomes across different model datasets(Table \ref{tab4}). Overall Survival (OS) dominates the outcome categories for all models, consistently representing around 64-70\% of the data. Objective Response Rate (ORR) follows as the second most represented category, ranging from 6\% to 11\%. Progression-Free Survival (PFS) is included in a smaller proportion, around 3-4\% across models. Safety indicators (AE and SAE) occupy a small portion, varying between 1.6\% and 4.9\%. Lastly, the "Other" category (for example: Relapse-Free Survival, Pharmacokinetic Parameters, Patient-Reported Outcomes) accounts for around 14-19\% of the outcomes across the datasets. Most clinical trials prioritize Overall Survival (OS) as the primary endpoint, and model performance on OS closely mirrors their balanced accuracy and MCC values.
\newline
Despite the small number of classifications for each clinical trial endpoint, GPT-4mini did not demonstrate the ability to predict clinical trial outcomes in any category. The same results were observed in GPT-3.5. Both models showed balanced accuracy around 0.5 and MCC values all below 0.1. Furthermore, the recall for these two models across all categories was close to 1, with specificity near 0, indicating that they tended to classify all clinical trial outcomes as "success" across all clinical trial endpoint categories.
\newline
For other models, performance across different clinical trial endpoint categories showed similar trends. For instance, among all models that exhibited some recognition capability, the ability to identify clinical trials with Overall Survival (OS) as the endpoint was the strongest. For GPT-4o and GPT-4, balanced accuracy in this category was significantly above the average (0.538 and 0.523). This phenomenon was also observed in Llama3. Safety indicators (AE and SAE) also demonstrated recognition capabilities in these models. Although performance in this category was not as strong as in the OS category, metrics for GPT-4, GPT-4o, and Llama3 were better than random guess. In particular, GPT-4o's specificity in this category reached 0.5, marking the highest level achieved by large language models in this experiment. However, the HINT model showed poor predictive performance for clinical trials focused on Overall Survival and safety indicators, with average accuracies of only 0.496 and 0.498, and MCC values of -0.028 and -0.007, respectively. This contrasts with the performance of the large models.
\newline
For other clinical trial classifications, such as Objective Response Rate (ORR), Progression-Free Survival (PFS), and other uncategorized endpoints, most of these models did not exhibit classification capabilities. Metrics for these models remained at random guess levels. Notably, for GPT-4o, which typically performed well, the MCC for ORR was -0.167, even falling below 0. In contrast, the HINT model showed relatively better predictive performance on these two categories.
\newline
Nonetheless, the specificity of the HINT model remained much higher than that of the large language models. This suggests that HINT maintained a relatively strong ability to recognize negative cases in these categories.
\begin{table}[h!]
\centering
\caption{Model Performance on Various Endpoints}\label{tab4}
\resizebox{\textwidth}{!}{
\begin{tabular}{@{}llllllll@{}}
\toprule
\textbf{Endpoint} & \textbf{Trials Number} & \textbf{Balanced\_acc} & \textbf{MCC} & \textbf{Recall} & \textbf{Specificity} \\ 
\midrule
\multicolumn{6}{c}{\textbf{Model: GPT-4mini}} \\ 
Overall Survival (OS) & 28 & 0.500 & 0.0 & 1.000 & 0.0 \\ 
Objective Response Rate (ORR) & 10 & 0.500 & 0.0 & 1.000 & 0.0 \\ 
Progression-Free Survival (PFS) & 5 & 0.500 & 0.0 & 1.000 & 0.0 \\ 
Safety Indicators (AE and SAE) & 12 & 0.500 & 0.0 & 1.000 & 0.0 \\ 
Others & 13 & 0.000 & 0.0 & 1.000 & 0.0 \\ 
\midrule
\multicolumn{6}{c}{\textbf{Model: GPT-4o}} \\ 
Overall Survival (OS) & 28 & 0.614 & 0.278 & 1.000 & 0.333 \\ 
Objective Response Rate (ORR) & 10 & 0.438 & -0.167 & 0.968 & 0.438 \\ 
Progression-Free Survival (PFS) & 5 & 0.500 & 0.000 & 0.946 & 0.0 \\ 
Safety Indicators (AE and SAE) & 12 & 0.583 & 0.169 & 0.962 & 0.500 \\ 
Others & 13 & 0.517 & 0.000 & 0.962 & 0.158 \\ 
\midrule
\multicolumn{6}{c}{\textbf{Model: GPT-4}} \\ 
Overall Survival (OS) & 20 & 0.562 & 0.281 & 1.000 & 0.125 \\ 
Objective Response Rate (ORR) & 9 & 0.500 & 0.000 & 1.000 & 0.000 \\ 
Progression-Free Survival (PFS) & 5 & 0.500 & 0.000 & 1.000 & 0.0 \\ 
Safety Indicators (AE and SAE) & 9 & 0.600 & 0.316 & 1.000 & 0.200 \\ 
Others & 7 & 1.000 & 0.000 & 1.000 & 0.062 \\ 
\midrule
\multicolumn{6}{c}{\textbf{Model: GPT-3.5}} \\ 
Overall Survival (OS) & 1423 & 0.504 & 0.044 & 1.000 & 0.013 \\ 
Objective Response Rate (ORR) & 409 & 0.499 & -0.023 & 0.998 & 0.000 \\ 
Progression-Free Survival (PFS) & 163 & 0.506 & 0.040 & 0.998 & 0.029 \\ 
Safety Indicators (AE and SAE) & 614 & 0.500 & -0.007 & 0.998 & 0.004 \\ 
Others & 463 & 0.504 & 0.000 & 0.998 & 0.012 \\ 
\midrule
\multicolumn{6}{c}{\textbf{Model: HINT}} \\ 
Overall Survival (OS) & 252 & 0.548 & 0.085 & 0.596 & 0.515 \\ 
Objective Response Rate (ORR) & 74 & 0.488 & -0.022 & 0.616 & 0.565 \\ 
Progression-Free Survival (PFS) & 38 & 0.446 & -0.120 & 0.619 & 0.692 \\ 
Safety Indicators (AE and SAE) & 89 & 0.516 & 0.034 & 0.605 & 0.366 \\ 
Others & 67 & 0.511 & 0.000 & 0.605 & 0.566 \\ 
\midrule
\multicolumn{6}{c}{\textbf{Model: Llama3}} \\ 
Overall Survival (OS) & 205 & 0.545 & 0.152 & 0.935 & 0.133 \\ 
Objective Response Rate (ORR) & 61 & 0.507 & 0.018 & 0.952 & 0.111 \\ 
Progression-Free Survival (PFS) & 26 & 0.476 & -0.098 & 0.951 & 0.000 \\ 
Safety Indicators (AE and SAE) & 88 & 0.528 & 0.081 & 0.954 & 0.161 \\ 
Others & 45 & 0.507 & 0.000 & 0.954 & 0.062 \\ 
\bottomrule
\end{tabular}}
\end{table}
\subsection{Prediction on different clinical trial period}\label{subsec2}
We observe that for all models, the GPT-4mini and GPT-3.5 models showed poor (close to random) predictive ability in clinical trials of both short and long durations. In contrast, the GPT-4 and GPT-4o models generally demonstrated better predictive performance for medium- and long-term clinical trials. The GPT-4 model showed higher average accuracy (0.562) in trials lasting over 3000 days, but did not exhibit predictive power for shorter trials. GPT-4o showed optimal results for trials lasting between 1000 and 2000 days, but overall displayed an upward trend in performance. For the HINT and Llama3 models, results were similarly poor for trials lasting 1000-2000 days (average accuracies of 0.495 and 0.501, respectively), but both models showed good predictive ability for both short- and long-term trials.(Table \ref{tab5}).
\newline
When examining specificity, the HINT model shows an increasing trend over time, indicating an improved ability to recognize negative cases in longer trials. However, this comes at the cost of a decrease in recall. In contrast, the specificity trends of GPT-4, GPT-4o, and other large language models closely mirror the trends in their average accuracy, suggesting that the ability to recognize negative samples is closely tied to the overall predictive performance of these models.
\newline
This trend suggests that GPT-4 and GPT-4o models are more capable of predicting medium- and long-term clinical trials. In contrast, HINT and Llama3 perform better in predicting short-duration clinical trials. As the duration of the trial increases, the HINT model shows improved ability to recognize negative samples but at the expense of a decrease in its ability to identify positive samples. The other two more basic GPT models still do not demonstrate significant predictive capability for clinical trials.
\newline
\begin{table}[h]
\centering
\caption{Model Performance on Clinical Trials with Different Durations}\label{tab5}
\begin{tabular}{@{}lllll@{}}
\toprule
\textbf{Trial Period} & \textbf{Balanced Accuracy} & \textbf{MCC} & \textbf{Recall} & \textbf{Specificity} \\ 
\midrule
\multicolumn{5}{c}{\textbf{Model: GPT-4mini}} \\ 
short-term trials & 0.500 & 0.000 & 1.000 & 0.000 \\ 
medium-length trials & 0.500 & 0.000 & 1.000 & 0.000 \\ 
long-term trials & 0.500 & 0.000 & 1.000 & 0.000 \\ 
\midrule
\multicolumn{5}{c}{\textbf{Model: GPT-4o}} \\ 
short-term trials & 0.417 & -0.218 & 0.833 & 0.000 \\ 
medium-length trials & 0.575 & 0.200 & 0.900 & 0.250 \\ 
long-term trials & 0.550 & 0.209 & 1.000 & 0.100 \\ 
\midrule
\multicolumn{5}{c}{\textbf{Model: GPT-4}} \\ 
short-term trials & 0.500 & 0.000 & 1.000 & 0.000 \\ 
medium-length trials & 0.500 & 0.000 & 1.000 & 0.000 \\ 
long-term trials & 0.562 & 0.240 & 1.000 & 0.125 \\ 
\midrule
\multicolumn{5}{c}{\textbf{Model: GPT-3.5}} \\ 
short-term trials & 0.504 & 0.058 & 0.998 & 0.011 \\ 
medium-length trials & 0.503 & 0.043 & 0.998 & 0.008 \\ 
long-term trials & 0.500 & 0.000 & 1.000 & 0.000 \\ 
\midrule
\multicolumn{5}{c}{\textbf{Model: HINT}} \\ 
short-term trials & 0.546 & 0.097 & 0.699 & 0.394 \\ 
medium-length trials & 0.495 & -0.011 & 0.505 & 0.485 \\ 
long-term trials & 0.524 & 0.049 & 0.364 & 0.684 \\ 
\midrule
\multicolumn{5}{c}{\textbf{Model: Llama3}} \\ 
short-term trials & 0.560 & 0.248 & 0.985 & 0.137 \\ 
medium-length trials & 0.501 & 0.007 & 0.951 & 0.052 \\ 
long-term trials & 0.566 & 0.219 & 0.959 & 0.176 \\ 
\bottomrule
\end{tabular}
\end{table}
\newline
\subsection{Terminated trials and prediction of termination}\label{subsec2}
Large language models (such as the GPT series and Llama3) performed poorly in predicting outcomes for terminated trials, with accuracy around 0.1 or lower, indicating their difficulty in handling the complex external factors causing trial termination(Table \ref{tab6}). Consequently, we excluded “terminated” trials from further analyses to avoid biasing the experimental results. While large language models could not identify trial termination, the HINT model demonstrated this ability, with stable accuracy across all clinical trial statuses, similar to its performance on the entire dataset, showcasing its effectiveness in distinguishing terminated trials.
\newline
\begin{table}[h]
\centering
\caption{Model Performance on Terminated Trials}\label{tab6}
\begin{tabular}{@{}lllllll@{}}
\toprule
\textbf{Model} & \textbf{HINT} & \textbf{GPT-3.5} & \textbf{GPT-4o} & \textbf{GPT-4mini} & \textbf{GPT-4} & \textbf{Llama3} \\ 
\midrule
Accuracy & 0.614 & 0.032 & 0.110 & 0.021 & 0.040 & 0.067 \\ 
\bottomrule
\end{tabular}
\end{table}
\newline
\FloatBarrier
\section{Methods}\label{sec3}
\subsection{Clinical Trials Data Curation and Processing}\label{subsec2}
The clinical trial data used in this study were sourced from the \textbf{ClinicalTrials.gov} database\cite{bib16}. We selected trials based on the following criteria\cite{bib17}:
\begin{itemize}
    \item \textbf{Population:} all genders and all age groups.
    \item \textbf{Study Phase:} limited to Phase 1, Phase 2, and Phase 3 trials.
    \item \textbf{Study Type:} interventional studies only.
    \item \textbf{Availability of results:} trials with available results data.
\end{itemize}
Additionally, we restricted the primary completion date of the trials according to the cut-off dates corresponding to the training periods of different models:
\begin{itemize}
    \item \textbf{GPT-3.5}~\cite{bib18}: trained till October 2021; we acquired 3,811 valid clinical trial records.
    \item \textbf{HINT}~\cite{bib9}: training checkpoint as of April 2022; we obtained a total of 2,304 records.
    \item \textbf{Llama3}~\cite{bib19}: trained till March 2023; we collected 725 records.
    \item \textbf{GPT-4}~\cite{bib20} and \textbf{GPT-4 Mini}~\cite{bib21}/\textbf{GPT-4o}~\cite{bib22}: trained till October 2023; we acquired 54 and 74 records, respectively (Fig.\ref{fig2}A).
\end{itemize}
Regarding trial status, we excluded categories with a very low number of records, such as \textit{"RECRUITING"} (only two records) and \textit{"SUSPENDED"} (only three records). The final retained trial statuses included \textit{"COMPLETED,"} \textit{"TERMINATED,"} and \textit{"ACTIVE\_NOT\_RECRUITING."} The distribution of trial statuses indicated that the \textit{"COMPLETED"} trials constituted the largest portion, dominating the dataset used (Fig\ref{fig2}B).
\newline
Through this filtering process, we extracted key information for each trial, including the title, brief summary, primary outcome measures, and study status for subsequent analysis. These data covered several critical aspects, including baseline characteristics, adverse events, participant flow, outcome measures, and supplementary information such as protocols and contact details. 
\newline
For most of our analyses, we retained clinical trial data with the status 
\subsection{Model Selection}\label{subsec2}
In this study, we have chosen to analyze several popular LLMs, both open-source and closed-source, alongside the deep learning model HINT. The LLMs we selected include OpenAI’s closed-source models GPT-3.5, GPT-4, GPT-4o, and GPT-4mini, as well as the open-source model Llama3 (Llama 8B-Instruct). Following (Table \ref{tab7}) provides specific parameters and configurations for each model.
\subsection{Clinical Trials' Label}\label{subsec2}
\begin{figure}[H]
\centering
\includegraphics[width=0.9\textwidth]{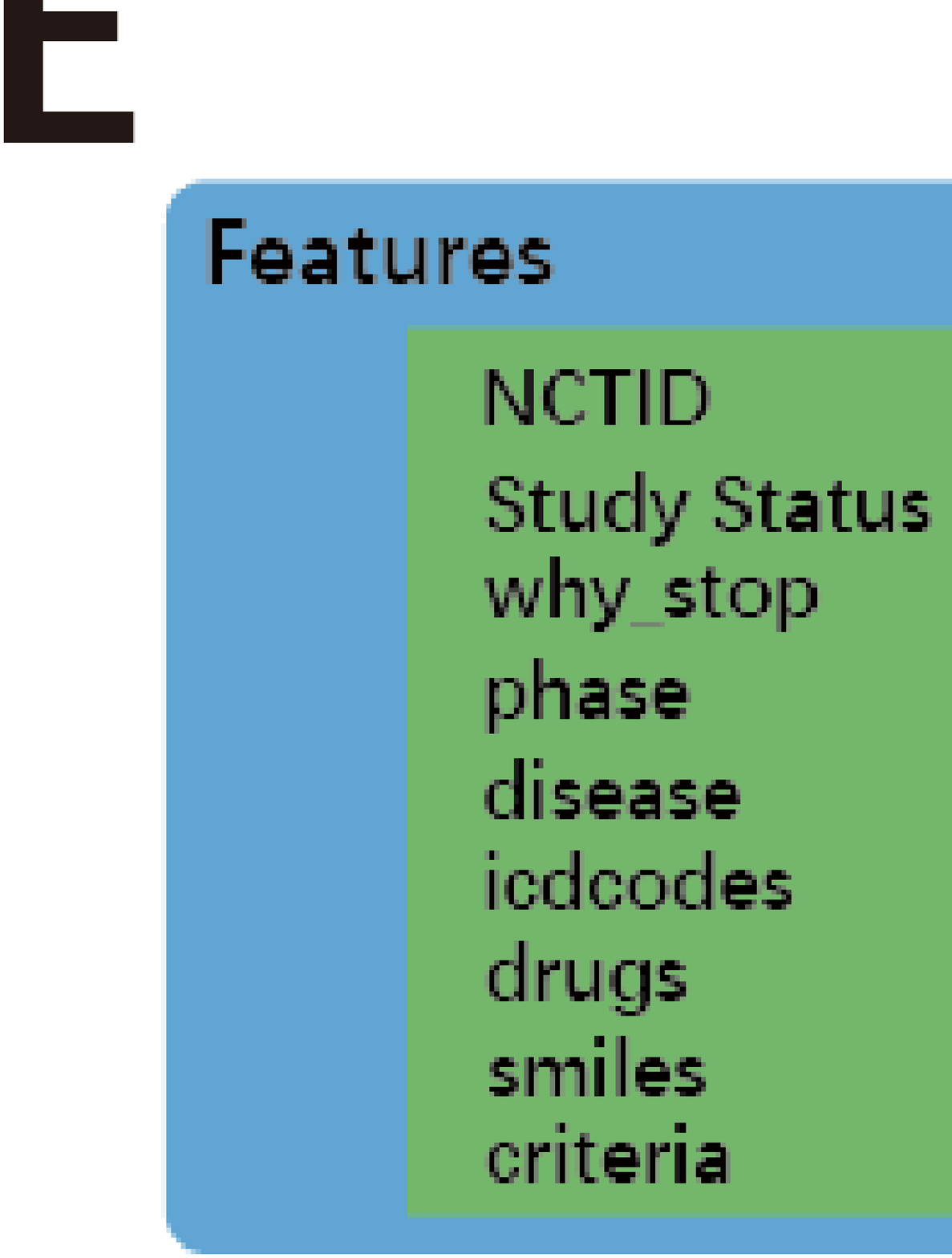}
\caption{A: The line chart shows the cumulative amount of clinical data covered by each model up to the respective cut-off dates, with the horizontal axis representing the time points of the models and the vertical axis representing the cumulative data volume. B: Data distribution across different clinical trial phases and trial statuses. C: Different labels in different study states: This figure describes the input data structures and output results for different models predicting clinical trial outcomes. The upper part shows the input features used by models. Based on these features, the models predict whether a trial is a "Success" or "Failure." E: This figure shows the input features used by the HINT model. The HINT model determines the final label by predicting the probability of success. F: Word cloud for clinical trials terminated reasons.}\label{fig2}
\end{figure}

The labeling process is carried out by two independent researchers using a double-blind cross-validation mechanism to ensure rigor. Discrepant cases are submitted for review by an independent panel consisting of clinical methodology experts and statisticians, who reach a final decision based on ICH\cite{bib23} guidelines and statistical consensus. The core categories for clinical trial failure include the lack of statistical significance for the primary endpoint efficacy, failure to meet predefined efficacy targets or breach of non-inferiority boundaries, insufficient sample size or missing evaluable data leading to inadequate statistical power, absence of synergistic benefits in combination therapies, flaws in treatment cycle design that prevent efficacy from being observed at key time points, mid-term ineffectiveness analysis triggering early termination of the study, severe safety issues or changes in drug regulatory status leading to trial interruption, overall shift in efficacy trends without reaching the significance threshold, and failure to identify a clinically meaningful effective dose range during the dose-finding phase.
\newline
In subsequent analyses, we excluded data with a " terminated " status because we found that the LLMs lacked the ability to accurately recognize trial termination. Fig\ref{fig2} C illustrates the proportion of trial status labels across various trial statuses within the entire dataset.
\newline
\subsection{Clinical Trials' Prediction}\label{subsec2}
During the prediction phase of the experiment, we applied several language models, including the GPT family, including GPT-3.5, GPT-4, GPT-4o, GPT-4mini and Llama3, to predict the outcomes of clinical trials. The GPT models are accessed via the OpenAI API, while the Llama3 model was deployed locally. The input context provided to the models included the trial title, summary, conditions, interventions, primary outcome measures, and trial type. We prompted the models with the following instruction: 
\newline
\newline
\textit{Use the following information to predict whether the outcome of the clinical trial will be a success or a failure. And only reply one word of the following: success or failure.}
\newline
\newline
We then parsed the response returned by the models to obtain the final prediction results.(Fig.\ref{fig2} D) 
\newline
For the HINT model, we input data in the required format\cite{bib9}, including the SMILES\cite{bib24} sequence information of the drug and the ICD-10 code of the studied disease(Fig.\ref{fig2} E). Following the preprocessing steps outlined by Fu et al.\cite{bib9}, we categorized the data based on the trial phase before feeding it into the HINT model for prediction. The model output is a probability of trial success; results with a success probability greater than 0.5 were labeled as “Success,” while those with a probability less than or equal to 0.5 were labeled as “Failure”.
\newline
\newline
\subsection{Keywords Extraction}\label{subsec2}
We conducted a keyword extraction analysis on the reasons for clinical trial termination and their primary endpoints. We utilized a network tool called Keyword Extractor\cite{bib25}, which is built on a language model and has demonstrated strong performance in keyword extraction tasks. By extracting keywords from these fields and calculating their frequencies, we aimed to identify common termination reasons and the primary focuses of the trials\cite{bib26}. This process helped us better understand the factors contributing to trial terminations and the main objectives of the studies.
\subsection{Metrics Selected}\label{subsec2}
We employed multiple evaluation metrics to analyze the performance of the models, including Balanced Accuracy,Matthews Correlation Coefficient, Recall, and Specificity:
\begin{itemize}
    \item \textbf{Balanced Accuracy:} The average of sensitivity and specificity, useful for class-imbalanced data. It reflects the model’s performance across both classes, offering a balanced view compared to traditional accuracy.
    \item \textbf{Matthews Correlation Coefficient (MCC):} A correlation measure using all categories in the confusion matrix. Ranging from -1 to +1, it’s ideal for imbalanced datasets, where +1 indicates perfect classification, 0 is random, and -1 is complete misclassification.
    \item \textbf{Recall (Sensitivity):} The proportion of true positives correctly identified amonPT all actual positives. This metric measures the model’s ability to capture relevant instances and avoid false negatives.
    \item \textbf{Specificity:} The proportion of true negatives correctly identified among all actual negatives. It indicates how well the model avoids false positives by correctly identifying negative cases.
\end{itemize}
These metrics allow for a comprehensive evaluation of the models’ predictive capabilities.
\begin{table}[h]
\centering
\caption{Model Versions and Training Cut-off Dates}\label{tab7}
\begin{tabular}{@{}lll@{}}
\toprule
\textbf{Name} & \textbf{Version} & \textbf{Training Cut-off Date} \\
\midrule
GPT-3.5 & gpt-3.5-turbo-0125 & Sep 2021 \\
GPT-4 & gpt-4-turbo-2024-04-09 & Dec 2023 \\
GPT-4o & chatgpt-4o-latest & Oct 2023 \\
GPT-4mini & gpt-4o-mini-2024-07-18 & Oct 2023 \\
Llama3  & Llama-3-8B-Instruct & March 2023 \\
HINT & HINT phase\_I/II/III.ckpt & Apr 2022 \\
\botrule
\end{tabular}
\end{table}
\subsection{Disease Classification}\label{subsec2}
For disease classification, we used ICD-10\cite{bib27} as the standard classification system, which is consistent with the method used during the training of the HINT model. 
\newline
We categorized these diseases into several groups using ICD-10 codes. These seven groups account for all trials in the baseline TOP and include:
\newline
\begin{itemize}
    \item \textbf{Respiratory Diseases:} Tuberculosis, sinusitis, tonsillitis.
    \item \textbf{Tumors/Cancers:} Cerebellar tumors, neuroectodermal tumors, breast cancer, gastric tumors.
    \item \textbf{Digestive System Diseases:} Dysentery, esophageal disorders, gastritis, duodenitis.
    \item \textbf{Nervous System Diseases:} Meningitis, Parkinson's disease, brain tumors.
    \item \textbf{Endocrine, Nutritional, and Metabolic Diseases:} Diabetes, hyperthyroidism, hyperlipidemia.
    \item \textbf{Non-pathological disease combinations:} Injuries, poisoning, and other external causes. External causes of morbidity and mortality. Factors influencing health status and healthcare utilization.
    \item \textbf{Other Uncategorized Conditions.}
\end{itemize}
\subsection{Clinical Trial Endpoint Grouping}\label{subsec2}
We also analyzed the predictive performance of trial outcome prediction across different endpoint subgroups. These four groups represent key areas in the overall trial analysis, with their respective accuracies:
\newline
\begin{itemize}
    \item \textbf{Overall Survival (OS):} Including survival rate, time to death, and mortality.
    \item \textbf{Objective Response Rate (ORR):} Such as response rate, tumor response rate, and overall response.
    \item \textbf{Progression-Free Survival (PFS):} Including progression-free time and time to disease progression.
    \item \textbf{Safety Indicators (AE and SAE):} Covering adverse events, serious adverse events, and toxic reactions.
    \item \textbf{Other uncategorized endpoints.}
\end{itemize}
\subsection{Clinical Trial Period Classification}\label{subsec2}
We also analyzed the impact of clinical trial duration on model performance by calculating the number of trial days based on the start and end dates, excluding any missing or outlier values. Due to limitations in the available dataset for the GPT-4 family models, we divided the clinical trial duration into three duration-based segments to examine whether varying lengths affect model predictive performance. This classification allows us to assess how different trial durations might influence the models’ predictive capabilities. Data statistics are shown in the following(Table \ref{tab8}).
\begin{itemize}
    \item \textbf{0-1000 days:} Some trials fall within this range, which we consider \textit{“short-term trials.”} (39.8\%)
    \item \textbf{1001-2000 days:} Most trials are within this range, which we classify as \textit{“medium-length trials.”} (50.4\%)
    \item \textbf{2001+ days:} A small portion of trials exceeds this duration, categorized as \textit{“long-term trials.”} (9.6\%)
\end{itemize}
\begin{table}[h!]
\centering
\caption{Data statistics}\label{tab8}
\resizebox{1\columnwidth}{!}{
\begin{tabular}{@{}lrrrrr@{}}
\toprule
\textbf{Category} & \textbf{Total} & \textbf{Completed} & \textbf{ACTIVE} & \textbf{Success} & \textbf{Failure} \\ 
\midrule
All Data & 2163 & 1802 & 358 & 1709 & 452 \\ 
0-1000 & 861 & 859 & 2 & 714 & 147 \\ 
1001-3000 & 1091 & 842 & 248 & 846 & 243 \\ 
3000+ & 207 & 101 & 104 & 145 & 62 \\ 
PHASE1 & 221 & 195 & 26 & 148 & 73 \\ 
PHASE2 & 1000 & 805 & 194 & 803 & 196 \\ 
PHASE3 & 636 & 537 & 98 & 555 & 81 \\ 
Diseases of the respiratory system & 148 & 125 & 23 & 116 & 32 \\ 
Neoplasms & 712 & 457 & 253 & 530 & 181 \\ 
Diseases of the digestive system & 192 & 167 & 25 & 156 & 35 \\ 
Diseases of the nervous system & 241 & 207 & 33 & 175 & 65 \\ 
Endocrine, nutritional and metabolic diseases & 173 & 162 & 11 & 147 & 26 \\ 
Non-pathological Diseases & 410 & 363 & 47 & 316 & 94 \\ 
Overall Survival (OS) & 1423 & 1201 & 220 & 1113 & 308 \\ 
Objective Response Rate (ORR) & 409 & 308 & 100 & 338 & 70 \\ 
Progression-Free Survival (PFS) & 163 & 91 & 72 & 128 & 35 \\ 
Safety Indicators (AE and SAE) & 614 & 550 & 63 & 380 & 233 \\ 
\bottomrule
\end{tabular}
}
\end{table}
\section{Conclusion}\label{sec4}
This study focuses on evaluating the performance of large language models (LLMs) in predicting clinical trial outcomes, with particular attention to Balanced Accuracy, Matthews Correlation Coefficient (MCC), Recall, and Specificity. Among all models, GPT-4o demonstrated the best overall performance, showing high balanced accuracy and MCC. However, similar to other LLMs, GPT-4o faced limitations in recognizing negative samples, a common challenge across these models. In comparison, the HINT model demonstrates consistently excellent ability to identify negative class samples.
\newline
In terms of negative sample recognition, GPT-3.5 and GPT-4mini exhibited almost no ability, severely limiting their utility in predicting failed trials. Llama3 showed some ability in this area, although it remained weak.
\newline
Different models show varying predictive abilities for clinical trials across different disease cateories. An important finding is that HINT struggles to predict the outcomes of oncology trials, which constitute the majority of the dataset. The complexity of cancer-related trials may be the reason for the model's underperformance in this area compared to large language models. Despite HINT demonstrating advantages in predicting certain outcomes and maintaining accuracy in identifying negative samples, its limitations in handling complex fields like oncology restrict its overall predictive capacity. Llama3 performs the best in predicting oncology clinical trial outcomes, and GPT-4o also shows corresponding capabilities. Different large language models excel in predicting outcomes for different types of diseases.
\newline
In contrast, the HINT model shows relatively stable performance across different trial phases and exhibits a clear advantage in identifying negative samples. As the trial phase progresses, the effectiveness of the HINT model continues to improve. Furthermore, HINT is resilient to external factors such as recruitment and funding issues. In contrast, the predictive ability of large language models is less influenced by the trial phase. GPT-4o and Llama3 perform exceptionally well in early phases (such as Phase I and Phase II), demonstrating strong predictive capability and robustness, complementing the limitations of the HINT model in these phases. However, HINT maintains a stronger ability to identify negative samples than all large language models.
\newline
Different models exhibit varying strengths in predicting specific clinical trial endpoints. Large language models, such as GPT-4o and GPT-4, show notable accuracy in predicting Overall Survival (OS), but struggle with more complex endpoints like Objective Response Rate (ORR) and Progression-Free Survival (PFS). Despite demonstrating strong performance in OS predictions, their ability to predict ORR and PFS often falls below random chance, indicating that they are more effective at handling simpler survival endpoints. In contrast, the HINT model demonstrates more consistent performance across various clinical trial endpoints and excels at recognizing negative samples, particularly in more complex categories such as ORR and PFS. While HINT's predictive performance is not as strong for certain outcomes like OS, its superior specificity and ability to identify negative cases make it a valuable model in clinical trial prediction.
\newline
The study also found that as trial duration increased, the predictive performance of the models generally improved for medium- and long-term clinical trials. For extended trials, the HINT model demonstrated enhanced ability to recognize negative samples, although this came at the expense of decreased recall. Meanwhile, GPT-4o and GPT-4 showed relatively stable performance, with GPT-4 achieving higher accuracy in trials lasting over 3000 days
\newline
In conclusion, GPT-4o and Llama3 stands out as the best large language model for clinical trial prediction, but all LLMs face challenges in recognizing negative outcomes.The performance of these two models varies in different scenarios and conditions, each with its own strengths and weaknesses. Nonetheless, these models are better suited for long-duration clinical trials. The HINT model’s high specificity in identifying negative outcomes highlights the need for further advancements in large language models to enhance their capability in negative predictions and in addressing external trial complexities.
\newline
\section{Authors' Contribution}\label{sec5}
Conceptualization: Lun Yu, Meijie Wang; Methodology: Shuyi Jin, Lun Yu; Data Curation and Labeling: Shuyi Jin, Lu Chen, Hongru Ding; Writing - original draft preparation: Shuyi Jin; Writing - review and editing: Lun Yu, Meijie Wang
\section{Declaration of Interests}\label{sec6}
Lun Yu and Meijie Wang are shareholders of Metanovas Biotech. Authors declare that they have no conflict of interest.
\section{Ethics Statement}\label{sec7}
This study does not contain any studies with human participants or animals performed by any of the authors.
\section{Data and Code Availability Statement}\label{sec8}
The dataset and code used in this study are available from the corresponding author upon reasonable request. 
\bibliography{bib}

\end{document}